\documentclass{article}
\usepackage{spconf,amsmath,graphicx}

\usepackage{enumitem}
\usepackage{anyfontsize}
\usepackage{amssymb}
\usepackage{amsmath}
\usepackage{tabularx}
\usepackage{bm}
\setlist{nosep, leftmargin=14pt}

\usepackage{mwe} 

\newcommand{\thisfontsize}[1]{{#1 The \string #1'' font size is: \f@size pt\par}}

\title{SFB-net for cardiac segmentation: bridging the semantic gap with attention}
\name{Nicolas Portal\textsuperscript{1,2}, Nadjia Kachenoura\textsuperscript{2}, Thomas Dietenbeck\textsuperscript{2}, Catherine Achard\textsuperscript{1}} 
\address{\fontsize{10}{10}\selectfont 1 - Sorbonne Université, CNRS, INSERM, Institut des Systèmes Intelligents et de Robotique, F-75005 Paris, France\\
        \fontsize{10}{10}\selectfont 2 - Sorbonne Université, CNRS, INSERM, Laboratoire d'Imagerie Biomédicale, F-75006 Paris, France} 

\begin{document}
%
\maketitle
\begin{abstract}
In the past few years, deep learning algorithms have been widely used for cardiac image segmentation.
However, most of these architectures rely on convolutions that hardly model long-range dependencies, limiting their ability to extract contextual information.
In order to tackle this issue, this article introduces the Swin Filtering Block network (SFB-net) which takes advantage of both conventional and swin transformer layers.
The former are used to introduce spatial attention at the bottom of the network, while the latter are applied to focus on
high level semantically rich features between the encoder and decoder.
An average Dice score of 92.4 was achieved on the ACDC dataset. To the best of our knowledge, this result outperforms any other work on this dataset.
The average Dice score of 87.99 obtained on the M\&M's dataset demonstrates that the proposed method generalizes well to data from different vendors
and centres.
\end{abstract}
\section{Introduction}
\label{sec:intro}

\subsection{Motivation}
Segmentation of cardiac structures on medical images gives crucial information 
to diagnose cardiomyopathies as it allows to accurately delineate structures targeted by the disease and monitor their remodelling. For example, inherited or acquired cardiomyopathies can be more easily diagnosed by doctors
assisted by effective segmentation software. Magnetic resonance imaging (MRI) is useful
to detect such diseases as it provides high contrast, resolution, and anatomical coverage, all without radiation and thus with low risk for patients.
Nowadays, many segmentation algorithms rely on deep learning methods as they achieved
good results on computer vision tasks.
The U-net architecture \cite{ronneberger2015u} is often used in practice as it proved to be an effective design to perform semantic segmentation. 
This architecture effectively fuses high-resolution information carried by the encoder with semantically-rich features of the decoder.
However, the simple concatenation of feature maps coming from the encoder and the decoder has been shown to be 
suboptimal \cite{zhou2018unet++}.
Indeed, feature maps coming from the encoder, though containing very fine-grained local details, carry less semantically rich features than
feature maps of the decoder. As a result they present a non-negligible amount of noise, but cannot be ignored because they
contain information on details, with a high spatial precision. This phenomenon is known as the \textit{semantic gap}.
Accordingly, this work focuses on designing a new mechanism able to bridge this semantic gap by filtering out noise found in the encoder
feature maps and focusing training on regions with strong response in the semantically-rich feature maps of the decoder.
\subsection{Related work}
Ronneberger et al.\cite{ronneberger2015u} introduced the U-net architecture which consists of an encoder to aggregate contextual information, 
a symmetric decoder to enable precise localization, and skip connections between the encoder and the decoder to exploit local information
contained in high-resolution feature maps.

Recently, attention mechanisms have been extensively used to improve neural networks performance.
\cite{fu2019dual} integrated both spatial and channel attention at the end of
the network to benefit from global contextual information.
Following \cite{dosovitskiy2020image} who showed that the multi head self-attention mechanism
used by transformers has the ability to increase the receptive field of the network
even at shallow layers, \cite{chen2021transunet}
used these transformer blocks at the bottom of the U-net network either in a 2D or 3D configuration
improving the performance of medical image segmentation algorithms.
Differently, \cite{gao2021utnet} use convolutions to reduce the spatial resolution of
feature maps, to subsequently incorporate self-attention at higher levels in the encoder.
\cite{liu2021swin} introduced a window and shifted window attention mechanism which improves performance over traditional
transformer blocks while reducing the computational complexity. As a result, these blocks
have been applied successfully to both 2D \cite{cao2021swin} and 3D \cite{hatamizadeh2022swin} medical image segmentation.
Attention has also been applied in the skip connection paths of the U-net,
first using summation \cite{oktay2018attention}, and more recently through
cross-attention mechanisms, either channel-wise \cite{wang2022uctransnet} or spatially, as a
way to filter out noise and focus on semantically rich features from deeper layers \cite{petit2021u}.

However, most methods dedicated to semantic gap reduction either use convolutions or self-attention. 
The few methods that use cross-attention either rely on entire transformer blocks \cite{peiris2021volumetric, wang2022uctransnet} 
or use full spatial attention \cite{petit2021u}, inducing high computational cost.

\vspace{-0.5em}
\section{Method}
\label{sec:method}
\subsection{A U-Net like architecture}
\begin{figure}[htb]
  \begin{minipage}[b]{1.0\linewidth}
    \centering
    \centerline{\includegraphics[width=8.5cm]{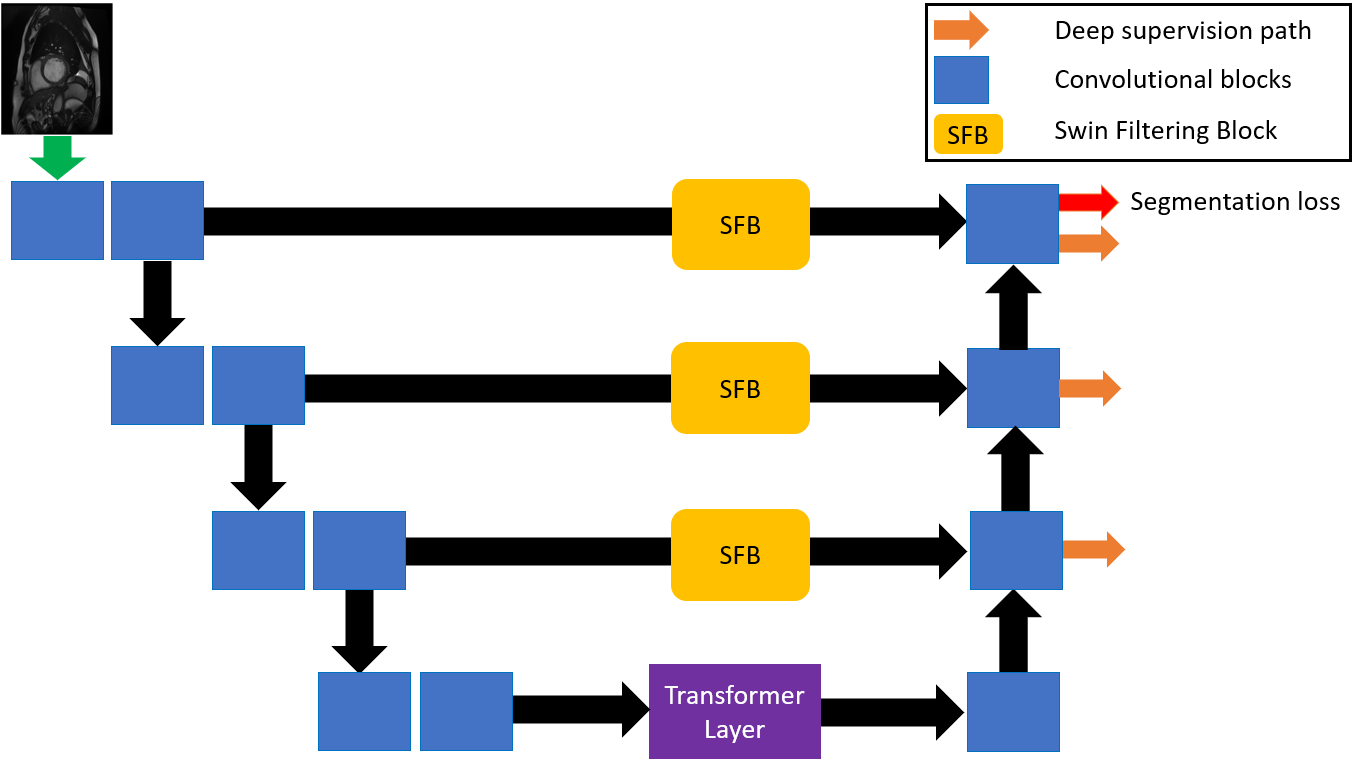}}
    \caption{Representation of SFB-net. Deep supervision is used in the decoder.
    Convolutional blocks are represented in blue, Swin Filtering Blocks (SFB) in yellow, and the conventional transformer layer in Purple.}
    \label{fig:model}
  \end{minipage}
\end{figure}

Our SFB-net is based on the U-net architecture \cite{ronneberger2015u} with an encoder, decoder, and skip-connections in-between, 
as illustrated in Figure \ref{fig:model}. Convolutional blocks, depicted in blue are used throughout the network. 
These blocks contain 2 convolutions, each followed by a batch normalization layer and a Gaussian Error Linear Unit (gelu) \cite{hendrycks2016gaussian} activation.
As proposed by \cite{myronenko20183d}, an encoder/decoder architecture is used where the number of convolutional blocks in the encoder is
doubled as compared to the decoder to improve the model encoding ability. The number of filters is doubled at each layer of the encoder and
halved for corresponding layers of the decoder.
Strided convolutions are used instead of pooling layers to down sample feature maps. The number of down sampling is limited resulting in a
feature map at the bottleneck 8 times smaller than the input image size. Up-sampling is carried out using 2D transposed convolutions.
To compensate for the resulting shallowness of the network, which may reduce its receptive field, a conventional transformer layer is introduced at the
bottleneck (depicted in purple). This enables the network to take advantage of global contextual information. Note that transformer
blocks are not used in the encoder and decoder since keeping convolutions at higher resolutions
was shown to give better results \cite{dai2021coatnet}. Indeed, convolutions generalize better to unseen images
than transformers and extract local information found at higher resolutions more effectively. 

Deep supervision is applied at each stage of the decoder.
More precisely ground truth segmentations are down sampled to match the size of the network's outputs. The loss weight $\alpha_i \forall i \in \{1, 2, 3\}$
for each resolution is halved when the image size is reduced. The final loss is the sum of successive stages loss and
is defined as:
\[\mathcal{L} = \alpha_1 \times \mathcal{L}_{{seg}_{\{H, W\}}} + \alpha_2 \times \mathcal{L}_{{seg}_{\{\frac{H}{2}, \frac{W}{2}\}}} + \alpha_3 \times \mathcal{L}_{{seg}_{\{\frac{H}{4}, \frac{W}{4}\}}}\]
For fair comparison with \cite{isensee2018nnu, zhou2021nnformer} a combination of cross-entropy and Dice loss is used to compute $\mathcal{L}_{seg}$.
Window-Multi Head Self Attention (W-MHSA) and Shifted-Window-Multi Head Self-Attention (SW-MHSA) \cite{liu2021swin}
are integrated between the encoder and the decoder in the skip connection paths to filter out noise
and highlight relevant information. This gating mechanism was introduced before the usual concatenation performed in the 
skip connection path of the U-net network. More details on these attention mechanisms, named Swin Filtering Block (SFB), 
are presented in the following section. SFB-net is shown in Figure \ref{fig:model}, with the SFB in yellow.

\subsection{Swin Filtering Blocks}
\begin{figure}[htb]
  \begin{minipage}[b]{1.0\linewidth}
    \centering
    \centerline{\includegraphics[width=8.5cm]{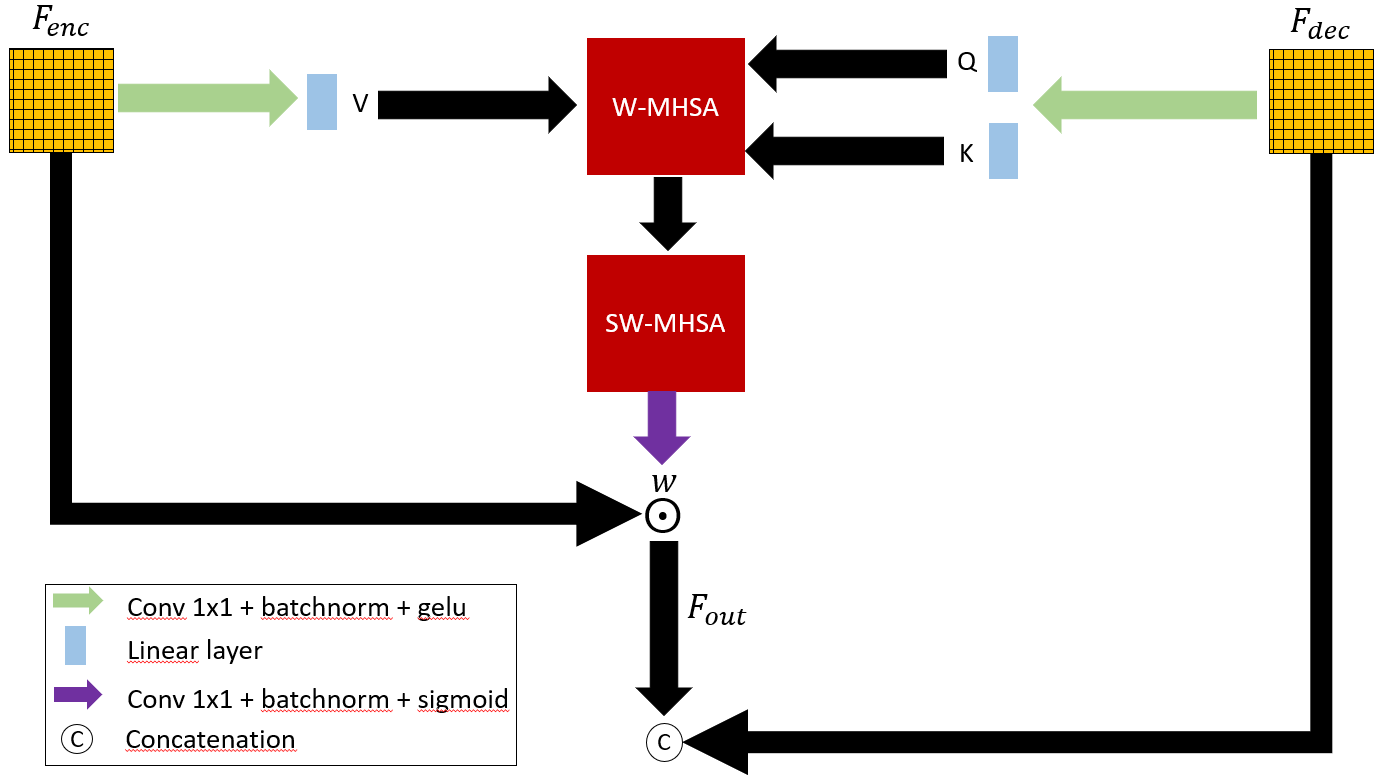}}
    \caption{Schematic representation of our Swin Filtering Blocks (SFB) used between the encoder and the decoder.}
    \vspace{-1.5em}
    \label{fig:sfb}
  \end{minipage}
\end{figure}

A filtering mechanism is introduced in the skip connection paths between the encoder and the decoder.
This process is described in Figure \ref{fig:sfb}.
The goal is to enable the decoder, to filter out irrelevant information originating from the encoder.
More precisely, the encoder's feature maps contain noise that should be discarded before concatenation with the decoder.
To do so, local information contained in high-resolution feature maps of the encoder that can be found in semantically-rich areas underlined
by the decoder are highlighted and emphasized, while response in other noisy areas are toned down. Similar to \cite{petit2021u}, Multi-Head-Self-Attention (MHSA) is used in this process.
However the window and shifted window version of MHSA introduced by \cite{liu2021swin} is preferred for its lower
computational load, reducing training time and enabling to use the saved GPU memory in other parts of the network.
Windowed Multi Head Self Attention (W-MHSA) performs attention in windows of $M$ by $M$ equal-sized patches. When performing Shifted 
Window Multi-Head Self Attention (SW-MHSA), windows are shifted by $\lfloor\frac{M}{2}\rfloor$ patches both in the x and y direction so that
attention can be conducted between patches belonging to different windows.
W-MHSA is described as:

\[\text{W-MHSA}(Q, K, V) = Softmax(\frac{QK^{T}}{\sqrt{d}} + B)V\]

Where Q, K and V $\in \mathbb{R}^{M^{2} \times d}$ are the query, key and value respectively. d is the query, key and value dimension. 
B $\in \mathbb{R}^{M^{2} \times M^{2}}$ is the learnable relative position bias added to each head which encode the spatial relationship
between patches. Q, K and V are tensors generated using separate linear layers. 
W-MHSA and SW-MHSA blocks are favorably used to perform cross-attention between the encoder and
the decoder's feature maps. Cross-attention uses the same process as self-attention but with key, query and value coming from different feature maps.
Since feature maps coming from the encoder are rescaled based on those of the decoder, 
values are chosen to come from the encoder while both query and key should come from the decoder:

\[\text{CA}_{out} = \text{SW-MHSA}(\text{W-MHSA}(Q_{F_{dec}}, K_{F_{dec}}, V_{F_{enc}}))\]

The result is passed to a sigmoid layer to generate weights $w$ ranging between 0 and 1 used to rescale the encoder feature map.

\[w = \sigma(\text{conv}(\text{CA}_{out}))\]

where $\sigma$ is the sigmoid function defined as $\sigma(x) = \frac{1}{1 + e^{-x}}$ and $conv$ is a standard convolution with 1 by 1 kernel
followed by batch normalization.
Finally the rescaled encoder feature map $F_{out}$ is obtained by applying the Hadamard product between computed weights $w$ and the original
encoder feature map:

\[F_{out} = F_{enc} \odot w\]

\section{Experiments}
\label{sec:experiments}
\subsection{Datasets}
Experiences are conducted on two datasets:
\begin{itemize}
  \item Automated cardiac diagnosis challenge dataset (ACDC) \cite{bernard2018deep}: this dataset comprises 100 patients corresponding to overall 1902 annotated slices. 
  Patients are divided into 5 groups according to specific diseases (normal, hypertrophic cardiomyopathies, dilated cardiomyopathies, 
  abnormal right ventricle and myocardial infarction). For each patient, right ventricular cavity (RV), myocardium (MYO)
  and left ventricular cavity (LV) at end systole and end diastole are labelled on slices covering the heart from its base to its apex. Only annotated slices are used in our experiments.
  A 5-fold cross validation is performed and dice score is used to compare SFB-net with literature.
  \item Multi-Centre, Multi-Vendor Multi-Disease Cardiac Image Segmentation Challenge (M\&Ms) \cite{campello2021multi}: this dataset comprises 375 patients corresponding to overall 7334 annotated slices.
  The dataset includes healthy subjects as well as patients with hypertrophic or dilated cardiomyopathies. Images were acquired in 4 different centres using 4 different MRI scanners.
  Thus, this dataset allows to test the generalizing ability of the proposed model.
  Similar to ACDC dataset, RV, MYO and LV are labelled at end systole and end diastole
  on all contiguous slices from base to apex. Data is already divided into training, validation and testing set. The 150 labelled subjects in the training set
  are used to train the network. Segmentation dice scores are reported on the available test and validation sets.
\end{itemize}
\subsection{Implementation details}
\label{sec:implementation}
SFB-net is implemented with Pytorch and trained using a 16GB Tesla v100 SXM2.
The nnUnet \cite{isensee2018nnu} framework is used as a starting point for this work. The AdamW optimizer and cosine annealing scheduler are
used for training. The initial learning rate and weight decay are both set to 0.0001. For fair comparison with
\cite{isensee2018nnu, zhou2021nnformer}, the number of training epochs is set to 1000. Each epoch is made up of 250 iterations. Batch size is 10 for ACDC and 6 for M\&M's.
Before training, all images are resampled in the x and y directions based
on the median pixel spacing of the dataset. 
As a post processing step, the largest connected component is kept in the binarized prediction (classes are merged). 
Please refer to \cite{isensee2018nnu} for additional details. 
A wide range of data augmentations is applied: rotation, scaling, gamma adjustment, brightness adjustment, mirroring, contrast modification,
low-resolution simulation, noise, and blur. Mirroring is also applied at testing time. The number of heads in SFBs are 2, 4 and 8. 
 The number of heads in the bottom transformer layer is set to 16. 
The maximum number of filters at the bottleneck of the network is 512. Image size is set to 224x224 pixels for ACDC and 288x288 pixels for M\&M's. SFB-net
is made up of 23 million parameters. 
In Table \ref{table:Ablationstudydice} GigaFLOPS (Gflops) are computed with a batch size of 1. 
The throughput is computed by averaging the time taken to process the largest possible batch size on the GPU over 100 iterations.

\subsection{Results}
\label{sec:results}
Comparison with literature on the ACDC dataset is summarized in Table \ref{table:ACDCresults} while comparison on the M\&M's dataset is
presented in Table \ref{table:MMsresults}.
Results for SFB-net and nn-Unet are obtained using a 5-fold cross-validation and reported after post-processing.
Other reported results come from their respective manuscripts. 
\begin{table}[ht]
\centering
\begin{tabularx}{\columnwidth}{|X|c|c|c|c|}
\hline
Methods & Mean & RV & MYO & LV \\
\hline
nnUnet \cite{isensee2018nnu} & 91.75 & 90.67 & 90.18 & 94.40 \\
$\Omega$-net \cite{vigneault2018omega} & 92.16 & \textbf{92.00} & 89.1 & 95.4\\
TransUnet \cite{chen2021transunet} & 89.71 & 88.86 & 84.54 & 95.73\\
SwinUnet \cite{cao2021swin} & 90.00 & 88.55 & 85.62 & \textbf{95.83}\\
UNETR \cite{hatamizadeh2022unetr} & 88.61 & 85.29 & 86.52 & 94.02\\
nnFormer \cite{zhou2021nnformer} & 92.06 & 90.94 & 89.58 & 95.65\\
\hline
SFB-net & \textbf{92.4} & 91.38 & \textbf{90.85} & 94.96\\
\hline
\end{tabularx}
\caption{Comparison with literature on the ACDC dataset. The metric used is dice score.}
\vspace{-0.5em}
\label{table:ACDCresults}
\end{table}

\begin{table}[ht]
  \centering
  \begin{tabularx}{\columnwidth}{|X|c|c|c|c|}
  \hline
  Methods & Mean & RV & MYO & LV \\
  \hline
  \cite{full2020studying} & \textbf{88.35} & \textbf{88.5} & \textbf{85.3} & \textbf{91.25} \\
  \cite{zhang2020semi} & 87.8 & 87.95 & 84.55 & 90.9\\
  \cite{ma2020histogram} & 87.35 & 87.5 & 84.05 & 90.5\\
  \cite{parreno2020deidentifying} & 87 & 85.75 & 84.1 & 91.15\\
  \cite{kong2020generalizable} & 86.57 & 86 & 83.3 & 90.4\\
  \cite{corral20202} & 86.62 & 86.3 & 83.35 & 90.2\\
  \hline
  SFB-net & 87.99 & 88.08 & 84.92 & 90.98\\
  \hline
  \end{tabularx}
  \caption{SFB-net results on the M\&M's dataset. 
          Dice scores of best performing algorithms in the M\&M's 2020 challenge are displayed for comparison.}
  \vspace{-0.5em}
  \label{table:MMsresults}
\end{table}

As compared to other work presented in Table \ref{table:ACDCresults}, we achieved the best overall Dice score (92.4\%),
as well as the best Dice score for the myocardium, while Dice scores of the LV and RV cavities were slightly lower. 
Of note, our average Dice score before post-processing was 92.31. 
When considering average Dice per patient, standard deviation is 0.033. 
Results on the ACDC dataset are better than those obtained on the M\&M's dataset (Table 2). This stems from the fact that M\&M's data comes 
from multiple vendors and centres whereas ACDC data was created by only one centre using a single brand MRI scanner. Nonetheless, as evidenced by
results in Table \ref{table:MMsresults},
our average Dice score ranks second in the 2020 M\&M's challenge revealing that our approach generalizes adequately to 
different vendors and centres.

Ablation study is conducted on the ACDC dataset to further assess the effectiveness of the proposed approach. 
Results reported below are those obtained before post-processing. The study is performed only on the first fold. 
Dice scores are reported in Table \ref{table:Ablationstudydice} for SFB-net as well as for its 2 variants:

\begin{itemize}
  \item no sfb: SFB-net without SFB.
  \item no trans: SFB-net with the transformer layer at the bottleneck replaced by one convolutional block.
\end{itemize}

Results shown in table \ref{table:Ablationstudydice} demonstrate that both SFBs and the transformer layer used at the bottom of the network contribute to increasing the overall model's performance.
However, while the addition of the transformer layer resulted in a higher Dice for all classes, SFBs did not bring any improvement for the myocardium
and the LV cavity.

\begin{table}[ht]
  \centering
  \small
  \begin{tabularx}{\columnwidth}{|X|c|c|c|c|c|c|}
    \hline
    Models & Mean & RV & MYO & LV & Gflops & FPS\\
    \hline
    SFB-net & 92.98 & 92.24 & 91.54 & 95.16 & 18.91 & 74.0\\
    no SFBs & 92.88 & 91.72 & 91.73 & 95.19 & 7.18 & 151.8\\
    no trans & 92.71 & 91.99 & 91.49 & 94.66 & 15.81 & 82.4\\
    \hline
    \end{tabularx}
    \caption{Ablation study conducted on the first fold of the ACDC dataset. Metric used is the Dice score. FPS refers to the throughput (image/s).}
    \vspace{-0.5em}
    \label{table:Ablationstudydice}
  \end{table}


\section{Conclusion}
\label{sec:conclusion}
This work introduced a new deep learning network architecture relying on spatial attention. Attention was applied both between the encoder and
the decoder as a filtering gate to focus learning on semantically-rich features, but also at the bottleneck of the network to increase the
receptive field and benefit from larger contextual information. Segmentation results of the left and right ventricles from MRI images of the ACDC and M\&M's challenges demonstrate the effectiveness and generalizability of the proposed approach. 

\vspace{-0.5em}
\section{Compliance with ethical standards}
This research study was conducted retrospectively using human subject data made available in open access. Ethical approval was not required as confirmed by the license attached with the open access data.
\vspace{-0.5em}

\section{Acknowledgments}
This work was funded by the grant number 965286 from the H2020 MAESTRIA project.



{\small
\bibliographystyle{IEEEbib}
\bibliography{strings,refs}
}
\end{document}